# On the Use of Skeletons when Learning in Bayesian Networks

Harald Steck*
Siemens AG, Corporate Technology
Neural Computation
Dept. Information and Communications
Otto-Hahn-Ring 6, 81730 Munich, Germany

## Abstract

In this paper, we present a heuristic operator which aims at simultaneously optimizing the orientations of all the edges in an intermediate Bayesian network structure during the search process. This is done by alternating between the space of directed acyclic graphs (DAGs) and the space of skeletons. The found orientations of the edges are based on a scoring function rather than on induced conditional independences. This operator can be used as an extension to commonly employed search strategies. It is evaluated in experiments with artificial and real-world data.

## 1 INTRODUCTION

Much progress has been made regarding structural learning in Bayesian networks. In the course of which, two main approaches have evolved. The one is constraint-based and relies on inducing conditional independences from the (empirical) probability distribution over the variables in a domain, e.g. [16], and the other tries to optimize a scoring function by means of a (heuristic) search strategy, e.g. [7, 9]. In the latter approach, scoring functions like the posterior probability or penalized likelihoods, e.g. the Bayesian Information Criterion (BIC), might be used. Finding the optimum Bayesian network structure was shown to be an NP-complete problem [5], so that one has to resort to heuristic search strategies.

Several Bayesian network structures, which are displayed by directed acyclic graphs (DAGs), can be Markov equivalent. An equivalence class contains all those DAGs which represent the same probability distributions determined by the conditional independences and dependences, see for instance [14].

*also: Dept. of Computer Science, Technical University of Munich, 80290 Munich, Germany; steck@in.tum.de

When learning from a probability distribution, score-equivalent scoring functions are commonly used, i.e. the same score is assigned to Markov equivalent DAGs. As a consequence, this can prevent local search procedures from finding optimum DAGs. For instance (cf. Fig. 1), assume that a local search procedure has arrived at the intermediate DAG (1). Let us further assume that the optimum DAG comprising the same edges is the one as depicted in (4). The DAGs (1), (2) and (3) are equivalent. Hence, only in the transition from DAG (3) to (4), there is a difference regarding the scores assigned to the DAGs. If exhaustive search is infeasible, the transitions from DAG (1) to DAG (3) via DAG (2) might not be found by a simple local search procedure, because these transitions are not related to an improvement regarding the score.

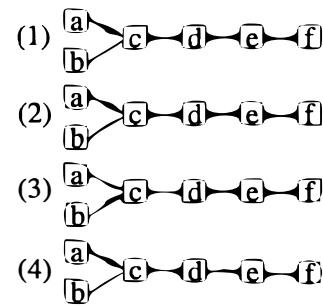

Figure 1: The DAG (1) is assumed to be the current one in a local search process, and DAG (4) is assumed to be the optimum one.

In order to get around this problem due to Markov equivalent DAGs, it has been proposed and examined to carry out local search in the space of equivalence classes rather than in the space of DAGs, which has several advantages, see for instance [12, 6]. A disadvantage is, however, that the number of neighboring graphs in the space of equivalence classes can be much larger than in the space of DAGs. Also, determining neighboring equivalence classes given an intermediate one can be computationally quite costly. Since this can render such an approach rather involved, it was proposed to alternate between the search space of DAGs and the space of equivalence classes for efficient computation [6].

In this paper, we present a heuristic search operator which makes use of skeletons. A skeleton is the graph obtained by substituting each directed edge in a DAG by an undirected one. The operator comprises two steps: Given an intermediate DAG in the search process, it is first transformed to its corresponding skeleton. In the second step, the proposed operator aims at orienting the edges in such a way that the resulting DAG is optimum (among the DAGs with the same skeleton) with respect to the used scoring function. In this paper, we focus on finding a single (optimum) graph as opposed to inducing several DAGs like it is done for model averaging, e.g. [17]. Since an equivalence class of DAGs is defined by its skeleton and the colliders (v-structures) [18], the heuristic search operator first identifies the colliders before orienting the remaining edges. This is similar to the constraint-based procedures used in the SGS or PC algorithms [16]. The main difference is that the presented operator orients the edges according to the used *scoring function* rather than according to the conditional independences and dependences induced from the (empirical) probability distribution. This rendered the presented operator quite robust in our computer experiments, whereas the edge-orientation procedures used in the SGS or PC algorithms are known to be rather unstable when applied to finite samples [16].

In the above simplistic example (cf. Fig. 1), the presented search operator yields the transition from the intermediate DAG (1) to the optimum DAG (4) at once without being affected by the equivalent DAGs (2) and (3). This is because it alternates from the DAG (1) to its skeleton, and then orients all edges again: First, the collider at the node $c$ is identified, which determines the equivalence class. Then the edges between $c$ and $d$, between $d$ and $e$, and between $e$ and $f$ are oriented such that no additional colliders occur. Alternating between the search space of DAGs and the space of skeletons might thus serve as an inexpensive alternative to using the search space of equivalence classes. The operator presented in this paper can help search strategies in the space of DAGs get out of local maxima, and might thus be used as an extension to commonly employed constraint-based approaches as well as to many greedy search strategies like local search.

Having applied this operator to an intermediate DAG during the search process, the orientations of *several* edges might be changed compared to the previous DAG. This is in contrast to local search, where subsequent DAGs differ from each other by exactly one directed edge, which is included, eliminated or reversed. The presented operator might thus be considered as non-local.

This paper is organized as follows. First, we are concerned with decomposable scoring functions, which are an essential requirement of the presented operator. For ease of computation, we approximate the posterior probability by employing the Bayesian Information Criterion (BIC). The next section provides the details of the operator: It focuses on the scoring function used for orienting the colliders (v-structures) and on the procedures for subsequently orienting the remaining edges. Finally, in our computer experiments, we used the operator as part of a very simple search strategy in order to compare the beneficial effects due to the presented operator to other search strategies like the K2 algorithm [7] or the local search strategy applied to learning in Bayesian networks [9].

## 2  DECOMPOSABLE SCORE

In this paper, we assume the scoring function to be *decomposable*, i.e. to factorize recursively like the joint probability distribution over the set of variables $\mathcal{V} = \{x_1, ..., x_n\}$ described by a Bayesian network,

$$p(x_1, ..., x_n) = \prod_{i=1}^{n} p(x_i \mid \text{pa}_m(x_i)), \quad (1)$$

where $\text{pa}_m(x_i)$ denotes the parents of the node[1] $x_i$ in the DAG $m$. Commonly used scoring functions factorize in this fashion in the case of complete data and in the absence of latent variables. As a particular choice for the scoring function, we use the posterior probability

$$p(m|D) = \frac{p(m)}{p(D)} p(D|m) \quad (2)$$

of the DAG $m$ when given data $D$. Simplifying the computation for practical reasons, we approximate the marginal likelihood $p(D|m)$ by the Bayesian Information Criterion (BIC),

$$\log p(D|m) \approx BIC(m) = \log L(\hat{\theta}) - \frac{1}{2} \log(N) |\hat{\theta}|. \quad (3)$$

The BIC states the trade-off between the maximum likelihood $L(\hat{\theta})$ of the model and its dimension $|\hat{\theta}|$ in an intuitive way. Unlike the Akaike criterion, the term penalizing model complexity depends on the number $N$ of cases in the given sample. The BIC is a rough approximation, and it avoids the introduction of priors over the parameters. For more details, the reader is referred to [11].

---

[1] For simplicity, we use *node* and *variable* synonymously.

First of all, let us consider two nested DAGs which differ from each other by exactly one edge. Let the corresponding two DAGs be denoted by $m_1$ and $m_2$, where the DAG $m_1$ contains one edge more than $m_2$, and let this additional edge in $m_1$ be oriented from node $a \in \mathcal{V}$ to node $b \in \mathcal{V} \setminus \{a\}$, indicated by $a \to b$. For brevity, we denote the parents of node $b$ in the sparser graph $m_2$ by $\pi = \text{pa}_{m_2}(b)$.

In order to compare the two DAGs $m_1$ and $m_2$ it is a natural choice to use posterior odds. With the above approximation, one arrives at the *relative* scoring function $G$ which gives the score of $m_1$ relative to $m_2$:

$$G(a, b, \pi) = \log \frac{p(m_1) \, BIC(m_1)}{p(m_2) \, BIC(m_2)}$$
$$= \log \frac{p(m_1)}{p(m_2)} + \log \frac{L(\hat{\theta}_1)}{L(\hat{\theta}_2)} - \frac{1}{2} \log(N) d_\text{f}$$

This score was used for model selection in [3], where the maximum *a posteriori* rather than the maximum likelihood configuration of the parameters was chosen. Assuming discrete variables with a multinomial distribution, the maximum likelihood ratio is given by

$$\log \frac{L(\hat{\theta}_1)}{L(\hat{\theta}_2)} = \sum_{a,b,\pi} N_{a\,b\,\pi} \log \frac{N_{a\,b\,\pi} N_{+ + \pi}}{N_{a+\pi} N_{+b\pi}}, \quad (4)$$

where the sum is over all configurations of $a$, $b$ and $\pi$. The counts from the data are denoted by $N_{a,b,\pi}$, where the "+" indicates that it is summed over the corresponding variable(s). Since the DAGs $m_1$ and $m_2$ differ in exactly one edge, the likelihood ratio depends only on the involved variables $a$, $b$, and $\pi$, and it is independent of the remaining ones. This holds also for the degrees of freedom $d_\text{f}$,

$$d_\text{f} = |\hat{\theta}_1| - |\hat{\theta}_2| = (S_a - 1)(S_b - 1) S_\pi, \quad (5)$$

where the number of (joint) states of a (set of) variable(s) is denoted by $S$. Instead of Eq. 5, one might use adjusted degrees of freedom [2].

In order to attain a prior $p(m)$ which decomposes in the same fashion, one may resort, for instance, to the assignment $p(m) \propto \exp(\alpha |\mathcal{E}(m)|)$ with the number of edges $|\mathcal{E}(m)|$ in the DAG $m$ and with a constant $\alpha$ for *a priori* penalizing DAGs dependent on the number of edges. Then the function $G$ depends only on the variables $a$, $b$ and $\pi$. It is score-equivalent [4].

If one is interested in a notably higher score of $m_1$ compared to $m_2$ one may use a threshold value $\gamma > 0$ in $G(a, b, \pi) \geq \gamma$, see for instance [11].

## 3 DETAILS OF THE OPERATOR

This section provides the details of the proposed search operator which alternates between the search space of DAGs and the space of skeletons. In the first step, the current DAG is transformed to its skeleton, which is simply done by dropping the directions of the edges. Given a skeleton in the second step, the aim is to find the orientations such that the resulting DAG corresponds to the maximum of the scoring function. Of course, this problem can only be tackled in a limited way. For simplicity, we use a greedy scheme, i.e. at an intermediate step of the edge-orientation process we orient the edge which increases the score most.

### 3.1 ORIENTING COLLIDERS

In this step, we use a partially directed acyclic graph (PDAG) which contains the same edges as the given skeleton. Initially, all its edges are undirected, at an intermediate stage some edge are oriented, and eventually all the edges of the PDAG are directed. In the PDAG, a structure which is a candidate for being a collider involves three nodes $a$, $b$ and $c$ such that there is an edge between $a$ and $b$ as well as between $b$ and $c$; there is no edge between $a$ and $c$. We denote an undirected edge between two variables $x$ and $y$ by $x \sim y$, an oriented one by $x \Rightarrow y$, and an edge with a proposed orientation by $x \to y$ (cf. Algo. 1).

In an intermediate PDAG $m_\text{int}$, it has to hold for a collider candidate that $a \hookrightarrow b \hookleftarrow c$ with $\hookleftarrow \in \{\sim, \Leftarrow\}$ and $\hookrightarrow \in \{\sim, \Rightarrow\}$. This means that the edges are not allowed to be already oriented like $a \Leftarrow b$ or $b \Rightarrow c$. For each of the collider candidates the score $G_\text{col}$ is calculated as described in the following section. Next, the collider candidate with the largest score $G_\text{col}(a, b, c \mid m_{int}) \geq \gamma$ is oriented like $a \Rightarrow b \Leftarrow c$. After a collider has been oriented, one has to make sure that there cannot occur a (directed) cycle in the PDAG when orienting one more edge. Such edges are thus oriented in the opposite direction without considering the scoring function for simplicity. After a collider or an edge have been oriented in the PDAG $m_\text{int}$, some of the scores $G_\text{col}(a, b, c \mid m_{int})$ have to be updated. The above steps are repeated while there are collider candidates with a score larger than $\gamma$.

### 3.2 SCORING COLLIDER CANDIDATES

Given an intermediate PDAG $m_\text{int}$, the score $G_\text{col}(a, b, c \mid m_{int})$ of a collider candidate $a \hookrightarrow b \hookleftarrow c$ is calculated by comparing DAGs with each other which are identical except of the orientations of the edges $a \hookrightarrow b$ and $b \hookleftarrow c$. In particular, the variables $a$, $b$ and $c$ have the same parents as in the PDAG $m_\text{int}$, i.e. $\text{pa}_{m_{int}}(a)$, $\text{pa}_{m_{int}}(b)$, and $\text{pa}_{m_{int}}(c)$, respectively. Let us denote the DAG with the collider $a \Rightarrow b \Leftarrow c$ by $m_\text{c}$ and the three alternative DAGs (without that collider) by

- $m_{\leftarrow}$ with the two edges oriented as $a \Leftarrow b \Leftarrow c$,
- $m_{\rightarrow}$ with the two edges oriented as $a \Rightarrow b \Rightarrow c$,
- $m_{\leftrightarrow}$ with the two edges oriented as $a \Leftarrow b \Rightarrow c$.

Like in Sec. 2, we use the approximation of the posterior odds in order to score the DAG $m_c$ relative to the alternative DAG $m_{\leftarrow}$ by the relative score $G_{\leftarrow}(a, b, c \mid m_{\text{int}})$,

$$\begin{aligned}
G_{\leftarrow}(a,b,c \mid m_{\text{int}}) &= \frac{p(m_c)\, BIC(m_c)}{p(m_{\leftarrow})\, BIC(m_{\leftarrow})} \\
&= G(a, b, \text{pa}_{m_{\text{int}}}(b) \cup \{c\}) \\
&\quad - G(a, b, \text{pa}_{m_{\text{int}}}(a)),
\end{aligned}$$

where we have used the scoring function $G$ as denoted in Eq. 4. The second line indicates a considerable simplification since the scoring function is decomposable. Analogously, one arrives at the scores of $m_c$ relative to $m_{\rightarrow}$, i.e. $G_{\rightarrow}(a, b, c \mid m_{\text{int}})$, and relative to $m_{\leftrightarrow}$, i.e. $G_{\leftrightarrow}(a, b, c \mid m_{\text{int}})$,

$$\begin{aligned}
G_{\rightarrow}(a,b,c \mid m_{\text{int}}) &= G_{\leftarrow}(c,b,a \mid m_{\text{int}}) \\
G_{\leftrightarrow}(a,b,c \mid m_{\text{int}}) &= G_{\leftarrow}(a,b,c | m_{\text{int}}) \\
&\quad + G(b,c,\text{pa}_{m_{\text{int}}}(b)) \\
&\quad - G(b,c,\text{pa}_{m_{\text{int}}}(c)) \\
&= G_{\rightarrow}(a,b,c | m_{\text{int}}) \\
&\quad + G(a,b,\text{pa}_{m_{\text{int}}}(b)) \\
&\quad - G(a,b,\text{pa}_{m_{\text{int}}}(a))
\end{aligned}$$

For reasons regarding the stability of the algorithm we use the following score of a collider candidate:

$$\begin{aligned}
G_{\text{col}}(a,b,c \mid m_{\text{int}}) = \\
\min\{G_{\leftarrow}(a,b,c \mid m_{\text{int}}), \\
G_{\rightarrow}(a,b,c \mid m_{\text{int}}), G_{\leftrightarrow}(a,b,c \mid m_{\text{int}})\}
\end{aligned} \quad (6)$$

If $G_{\text{col}}(a, b, c \mid m_{\text{int}}) > 0$, this score is a lower bound for the increase of the overall score of the graph which comprises a collider compared to an alternative graph which is identical with $m_c$ except of this collider. If $G_{\text{col}}(a, b, c \mid m_{\text{int}}) < 0$, this is a bound for the maximum decrease of the score when these edges are oriented as a collider compared to one of the alternative orientations.

### 3.3 ORIENTING REMAINING EDGES

After the colliders have been oriented, the algorithm orients the remaining edges as follows. Guided by the heuristic of parsimony the aim is to find an orientation of the remaining edges such that a minimum number of additional colliders occurs. In the ideal case, where a perfect map of the probability distribution over the variables exists, this is possible without introducing an additional collider. In this case, the procedure *ProposeOrientations* (cf. Algo. 1) indicates the orientations of the remaining edges by arrows of the type $\rightarrow$ or $\leftarrow$, and no arrows of the type $\leftrightarrow$ are found; edges whose orientations can differ in equivalent DAGs are indicated by $\sim$, and they are oriented in the final step.

---
**procedure** *ProposeOrientations*
**input:** PDAG $m_{int}$ with edges $\sim, \Rightarrow, \Leftarrow$.
**output:** PDAG $m_{int}$ with edges $\sim, \Rightarrow, \Leftarrow, \rightarrow, \leftarrow, \leftrightarrow$.
**(1)** $\forall a, b, c \in \mathcal{V}$: if $a \hookrightarrow b \hookleftarrow c$ with $\hookrightarrow \in \{\Rightarrow, \rightarrow, \leftrightarrow\}$ and $\hookleftarrow \in \{\sim, \leftarrow\}$ and no edge between $a$ and $c$ then substitute either $b \sim c$ by $b \rightarrow$ or $b \leftarrow c$ by $b \leftrightarrow c$.
**(2)** $\forall a, b \in \mathcal{V}$: while $\exists\ a \leftarrow b$ and $\exists\ a = x_1, ..., x_q = b\ (x_i \neq x_j, q > 2)$ such that $x_{i-1} \hookrightarrow x_i$ $(i = 2, ..., q)$ with $\hookrightarrow \in \{\Rightarrow, \rightarrow, \leftrightarrow\}$ then substitute $a \leftarrow b$ by $a \leftrightarrow b$.
**(3)** while edges can be oriented go to (1).

---
Algorithm 1: The procedure *ProposeOrientations* tries to orient the edges in such a way that no additional colliders occur. If this is not possible, edges are oriented in "both" directions, indicated by $\leftrightarrow$.

---
**procedure** *FixOrientations*
**input:** PDAG $m_{int}$, data $D$
**output:** PDAG $m_{int}$
**(1)** call *ProposeOrientations*.
**(2)** substitute uniquely oriented edges $a \rightarrow b$ by $a \Rightarrow b$.
**(3)** among all structures of the form $a \leftrightarrow b \leftrightarrow c$ without an edge between $a$ and $c$, orient the one with the highest score $G_{\text{col}}(a, b, c | m_{\text{int}})$ like $a \Rightarrow b \Leftarrow c$.
**(4)** if there are edges of the type $\leftrightarrow$ then substitute all those by $\sim$ and go to (1).

---
**(∗)** $\forall a, b \in \mathcal{V}$: while $\exists\ a \hookleftarrow b$ with $\hookleftarrow \in \{\sim, \leftarrow, \rightarrow, \leftrightarrow\}$ and $\exists\ a = x_1, ..., x_q = b\ (x_i \neq x_j, q > 2)$ such that $x_{i-1} \Rightarrow x_i\ (i = 2, ..., q)$ then orient $a \Rightarrow b$.

---
Algorithm 2: This procedure resolves inconsistencies regarding the orientations of edges (indicated by $\leftrightarrow$) by using the scoring function.

In step (2) of the procedure *FixOrientations* (cf. Algo. 2), the orientations of those edges are *fixed* (by a double-arrow $\Rightarrow$ or $\Leftarrow$) whose orientations are uniquely indicated by the procedure *ProposeOrientations* (denoted by $\leftarrow$ or $\rightarrow$). However, if an edge of the type $\leftrightarrow$ is *proposed* by the procedure *ProposeOrientations*, the edges cannot be oriented such that no additional colliders occur. Hence, in step (3) of the procedure *FixOrientations*, we propose to orient the pair of ambiguously oriented edges like a collider which gets the highest score. Note that the score $G_{\text{col}}$ might be negative in this case. Hence, orienting the highest scoring edges like a collider corresponds to decreasing the overall score as little as possible in this greedy scheme.

Like the procedure *ProposeOrientations*, also *FixOrientations* has to make sure that a (directed) cycle cannot occur when fixing the orientations of the edges. Hence, the step (∗) has to be carried out each time an edge has been oriented like ⇐ or ⇒ by the procedure *FixOrientations*. Note that (∗) focuses on possible cycles due to edges whose orientations are already *fixed*.

After *FixOrientations* is finished, there might still be undirected edges in the PDAG, as already mentioned. Those edges are oriented in a final step of the edge-orientation procedure. This is done by randomly picking one of those edges and by assigning to it a random orientation. Then the procedure *FixOrientations* is called in order to orient the edges which are affected by this assignment. This is repeated until all edges are oriented.

For illustration, let us consider the example in Fig. 2: Assume that the only collider with a score $G_{\text{col}} \geq \gamma$ is $b \Rightarrow c \Leftarrow d$, cf. (2). When orienting the remaining edges, *ProposeOrientations* indicates inconsistencies (↔) due to a possible directed cycle, cf. (3). Hence, an additional collider is introduced, namely the one with the highest score $G_{\text{col}} < \gamma$, e.g. at the node $f$, see (4). In subsequent steps, no inconsistencies occur, so that the orientation of the edge between $c$ and $g$ – and hence the equivalence class – can be identified, cf. (5) and (6).

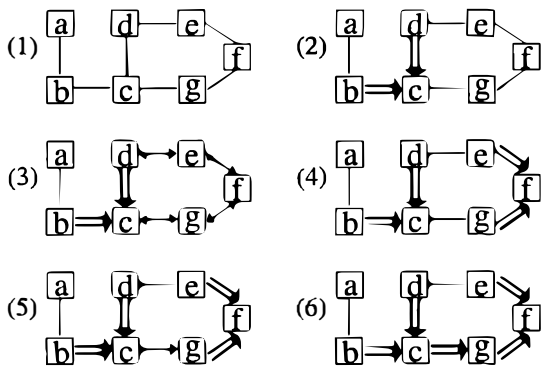

Figure 2: A simplistic example how the procedures *ProposeOrientations* and *FixOrientations* work.

### 3.4 STABILITY

The presented procedure for orienting the edges given a skeleton is similar to the ones used in the SGS and PC algorithms [16]. The main difference is that our procedure aims at finding the orientations such that a scoring function is maximized. Of course, since a greedy approach is taken, this procedure might only find a local optimum. In contrast, the SGS and PC algorithms orient the edges based solely on the conditional independences induced from the data. This is correct, if the probability distribution is perfectly known and if there exists a perfect map [16]. Given *finite* data, however, the latter procedures are known to be unstable (cf. [16]), whereas we found the presented procedure to be quite stable in our computer experiments.

### 3.5 LIMITATIONS OF THE HEURISTIC

Finding the skeleton corresponding to a given DAG is straight-forward. In contrast, the second step of the presented operator is, of course, not guaranteed to find the global optimum regarding the orientations of the edges, since learning Bayesian networks is an NP-complete problem [5]. However, it might serve as an efficient way of finding close to optimum orientations.

The heuristic operator calculates the score of the collider candidates based on the edges which have already been oriented in the *intermediate* graph. Of course, this can only be an approximation to the "correct" scores. In particular, at the beginning of this procedure, when only a small number of edges has already been oriented, the computed scores might considerably differ from the "correct" scores, because the optimum parents might not be known at this stage. Hence, this greedy procedure might get stuck at a local optimum. When orienting the remaining edges, there might occur inconsistencies regarding the directions, which indicate that additional colliders have to be introduced in order to attain a DAG. This is done in a heuristic way by orienting those edges like a collider which entail the smallest decrease regarding the score. Additionally, it has to be made sure that no directed cycles arise during this procedure. This is done in a simple heuristic way without considering the score. For an evaluation of the presented heuristic operator one has to resort to computer experiments.

## 4 PRELIMINARY EXPERIMENTS

In our computer experiments, we used the presented operator together with two additional operators, namely one for forward inclusion and one for backward elimination, in order to arrive at a simple search strategy, which we call *skeleton search strategy* in the remainder of this section. We compared this search strategy with other ones like the K2 algorithm [7] and the greedy algorithm based on local search, as described for instance in [9]. In detail, this simple search strategy works as follows: Starting out with the empty DAG, the graph is optimized by repeatedly cycling through the following three steps:

1. apply the presented operator

2. carry out one step of forward inclusion in the space of DAGs

3. perform backward elimination in the space of DAGs

After the orientations of the edges have been optimized by the presented operator, the edge which improves the score most is included into the DAG. Of course, no (directed) cycles may occur in the resulting graph. In this scheme, at most *one* edge is included into the graph in each cycle of the algorithm so that the presented operator can optimize the orientations before the next edge is included. The third step allows the algorithm to remove edges which have been included previously when the orientations of some edges might have been different. It might occur that this simple search strategy oscillates among two or more DAGs at the end of the search process, i.e. it may not converge to a unique DAG. This is because the presented operator is non-local in the sense that it possibly changes the orientations of more than one edge. For the stopping criterion this has to be taken into account, and we propose to keep track of the last few DAGs and to eventually choose the one with the highest score.

The results in our computer experiments were obtained from the following real-world data sets: The data (SEW) was gathered by Sewell and Shah [15], and it is concerned with high-school students. It comprises 5 variables and over 10,000 cases. The data (ENV) is concerned with environmental influences on the condition of trees [10]. It contains 11 variables and 6168 cases. In our experiments, we ignored the first and the ninth variable of this data set, since the latter is not documented and the former has several hundred states which encode the position of the trees with respect to a grid on a map. Moreover, we discretized the last two variables in this data set such that each variable had four states. We obtained the data (BOS) by discretizing the variables of the Boston Housing Data [8] such that each variable became binary. This sample contains 14 variables and 506 cases. The data (CAR) contains 46 variables, and 794 cases were available from [13]. It was gathered from customers of a car company. The two variables VORMO and URTEI have 16 and 10 states, respectively. For practical reasons, we summarized some of their states such that these two variables had 3 and 4 states, respectively.

When comparing the different search strategies we applied the same scoring function $G$ (cf. Eq. 4) to each of the strategies. We committed ourselves to a uniform prior over the DAGs, $p(m) = $ const, and imposed no constraints on the network structures. The threshold value $\gamma/2 = 3$ was used in order to include only those edges into the graph which lead to some notable increase in the score [11]. The parameters $\theta_m$ of an induced DAG $m$ were calculated in a Bayesian way, namely as the expectation values of the posterior distributions over the parameters. We used conjugate priors and a small equivalent sample size of $N' = 5$, see for instance [9].

Table 1: Experiments with four real-world data sets. The results are assessed by 5-fold cross validation (XV), cf. Sec. 4 for details.

| data set | nodes | skeleton search XV | local search XV |
|---|---|---|---|
| SEW | 5 | $0.0504 \pm 0.0035$ | $0.0514 \pm 0.0034$ |
| ENV | 11 | $0.877 \pm 0.018$ | $0.921 \pm 0.040$ |
| BOS | 14 | $1.94 \pm 0.11$ | $2.15 \pm 0.12$ |
| CAR | 46 | $25.53 \pm 0.26$ | $25.54 \pm 0.26$ |

In order to assess the expected utility of the Bayesian networks induced by the different search strategies we carried out 5-fold cross validation (XV), cf. [9]. The Kullback-Leibler divergence was used in order to compare the probability distributions underlying the validation sets with the ones described by the induced Bayesian networks. In Table 1, we depicted the mean and the standard deviation of the Kullback-Leibler divergence when averaged over the 5 samples in cross validation. Small values indicate good learning results. We compared the skeleton search strategy to the greedy algorithm based on local search as described, for instance, in [9].

Considering cross validation, the skeleton search strategy yields slightly better DAGs than does local search for each of the data sets in Table 1. The differences might be considered notable regarding the data sets (ENV) and (BOS), because the difference in the means is not considerably smaller than the standard deviations. This does not hold for the data sets (SEW) and (CAR). Considering the used scoring function $BIC$ in the experiment with the data (SEW), the simple skeleton search strategy found the global optimum, whereas local search found only a local optimum. The large values concerning the tiny data set (CAR) might indicate that this data set cannot be well described by means of a Bayesian network model.

In our Alarm network [1] experiments, we used the posterior probability with conjugate priors as the scoring function, see for instance [9]. The learning algorithms employed the corresponding relative score $G$ which can be derived for the posterior probability with conjugate priors analogously to Eq. 4. A small value of the equivalent sample size was chosen, i.e. $N' = 1$. We note that the Kullback-Leibler divergence computed in cross validation did not very sensitively depend on the chosen equivalent sample size which we varied between $N' = 1$ and $N' = 100$. This was also noted in [9]. Regarding the number of edges in the induced graph, we observed their number to increase when the equivalent sample size $N'$ rises.

In order to examine the stability of the presented operator concerning different sample sizes, we applied the skeleton search strategy to data sets of various sizes sampled from the probability distribution described by the Alarm network [1]. The Alarm network contains 37 discrete variables and 46 edges. Since the (partial) ordering of the variables is known, we could use the K2 search strategy [7] for comparison. The K2 procedure requires an initially specified ordering of the variables, and one can, hence, expect this search strategy to be quite stable regarding different sample sizes. In Fig. 3, we compare the skeleton search strategy to the K2 search strategy. Given only small samples (with less than 2,000 cases), our search strategy gets stuck at a local optimum which is far from the global one. The prior knowledge about the correct ordering of the variables seems to be crucial for the K2 procedure in the case of small samples. However, from samples containing at least 2,000 cases, the skeleton search strategy induces DAGs which get a higher score than the ones found by the K2 search strategy. Considering structural differences, we noticed that the K2 algorithm tends to include slightly more edges into the graph than the skeleton search strategy does. Compared with the original Alarm network structure, one edge was usually "erroneously" removed by both search strategies in most of our experiments, since this was favored by the scoring function. Except of this, the skeleton search strategy either induced the correct structure or – in particular when given small data sets – got stuck at a local optimum which erroneously contained an additional edge which was related to a "wrong" orientation of another edge. The K2 algorithm usually included one, and sometimes two, edges "erroneously".

Finally, we examined how quickly the skeleton search strategy reaches the end of the search process, i.e. how many cycles it takes to identify an approximately optimum structure. Since in each cycle at most one edge is added, the number of cycles cannot be smaller than the number of edges present in the induced DAG. In our Alarm network experiments, we found that the number of edges being present in the graph increases quite quickly before the end of the search process is reached (cf. Fig. 4). Since information about the ordering of the variables is not given as input to our algorithm, edges are "erroneously" included into the graph. Towards the end of the search process, after the orientations of the edges have been established, most of the "erroneously" included edges are removed in this example. We found that the resulting DAG got a higher score than the one induced by the K2 search strategy, although the latter used the correct ordering of the variables as prior knowledge.

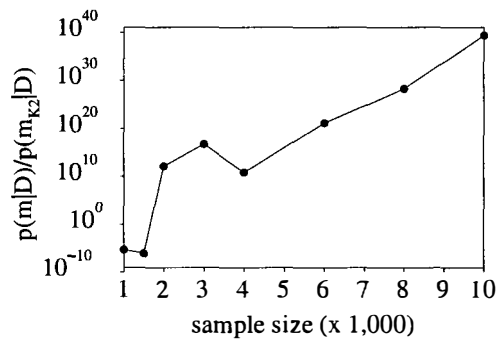

Figure 3: In our Alarm network experiments, the posterior probability of the DAG $m$ induced by the skeleton search strategy is compared with the DAG $m_{K2}$ which results from the K2 search strategy. The geometric mean over 5 samples for each size is depicted.

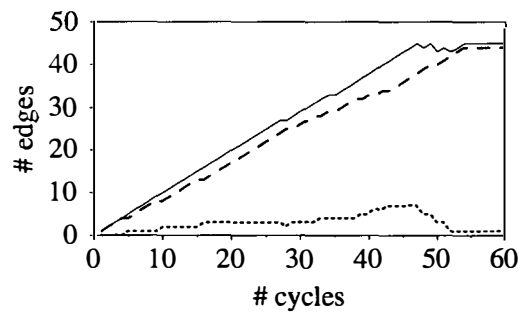

Figure 4: Time evolution of the number of edges during the learning process in one of our experiments with the Alarm network using a sample with 2,000 cases. The solid line indicates the overall number of edges being present in the intermediate graph $m_{\text{int}}$ after each cycle; it is subdivided into the edges which are also present in the original Alarm network (dashed line) and the ones which are not (dotted line).

In our computer experiments, we observed that computing the counts $N$ of the different configurations of several variables is typically very time consuming. In fact, most of the computation time was spent on this task. Hence, a lot of computation time can be saved by caching the scores $G$ once they have been computed. Since the DAGs in subsequent cycles are usually similar to each other, this caching can lead to a large speed-up.

## 5 CONCLUSIONS

In this paper, we presented a search operator which makes use of skeletons. In order to derive a DAG when given a skeleton, the edges are oriented with the aim of finding the optimum orientations with respect to a scoring function. This is in contrast to other procedures of this kind, e.g. the ones used in the PC or SGS algorithms [16], which are based on induced conditional independences, and which are known to be unstable when applied to finite data. Since the presented operator might simultaneously change the orientations of several edges it can be considered as non-local in the space of DAGs. The presented operator first identifies the colliders, which determine the equivalence class, and subsequently orients the remaining edges. This operator was thus quite robust against problems entailed by Markov-equivalent network-structures in our computer experiments. The operator was used in a simple search strategy which we compared to commonly used learning algorithms. Our preliminary experiments with artificial and real-world data suggest that alternating between the search space of DAGs and the space of skeletons might serve as a powerful and computationally efficient alternative to more involved search strategies.

**Acknowledgments**

I would like to thank Steffen Lauritzen for his hospitality and for numerous enlightening discussions. I am also grateful to Paul Theo Pilgram, Volker Tresp, and anonymous reviewers for valuable comments which helped improve this paper. This work was supported by an Ernst-von-Siemens grant.